\pdfoutput=1

\documentclass[11pt]{article}

\usepackage[]{EACL2023}
\usepackage{times}
\usepackage{latexsym}
\usepackage{tabulary}
\usepackage{graphicx}
\usepackage{bbm}

\usepackage[normalem]{ulem}
\useunder{\uline}{\ul}{}
\usepackage{xspace}
\usepackage{sidecap}
\usepackage{booktabs, multirow}
\usepackage{array}

\usepackage{longtable}
\usepackage{times}
\usepackage{latexsym}

\newcolumntype{H}{>{\setbox0=\hbox\bgroup}c<{\egroup}@{}}
\usepackage[T1]{fontenc}

\usepackage[utf8]{inputenc}

\usepackage{microtype}
\usepackage{inconsolata}
\newcommand{\nameExpOne}{climabench classification\xspace}
\newcommand{\nameExpTwo}{in-domain self-supervised questionnaire filling\xspace}
\newcommand{\nameExpThree}{cross-domain questionnaire filling\xspace}
\newcommand{\nameExpFour}{unstructured questionnaire filling\xspace}

\newcommand{\nameTypeStakeholder}{\textit{stakeholder-type}\xspace}
\newcommand{\nameStakeholder}{\textit{stakeholder}\xspace}

\newcommand{\nameSemiDoc}{semi-structured document\xspace}

\newcommand{\nameCDPDataset}{\textsc{CDP dataset}\xspace}
\newcommand{\varCDPDataset}{D_{cdp}\xspace}
\newcommand{\varCDPCities}{D_{city}\xspace}
\newcommand{\varCDPCompanies}{D_{corp}\xspace}
\newcommand{\varCDPStates}{D_{state}\xspace}

\newcommand{\varQuestions}{\textit{$Q$}\xspace}
\newcommand{\varQuestion}{q\xspace}
\newcommand{\varAnswers}{\textit{$A$}\xspace}
\newcommand{\varAnswer}{a\xspace}

\newcommand{\climaCDPQAClassification}{\textsc{CDP-QA}\xspace}
\newcommand{\climaCDPTopicClassification}{\textsc{CDP-Topic}\xspace}

\newcommand{\climabench}{\textsc{ClimaBench}\xspace}

\newcommand{\CAP}{\textsc{CAP}\xspace}
\newcommand{\claps}{\textsc{CAPs}\xspace}
\newcommand{\climaCDP}{\textsc{Clima-CDP}\xspace}
\newcommand{\climaQA}{\textsc{Clima-QA}\xspace}

\newcommand{\climaInsurance}{\textsc{Clima-Ins}\xspace}
\newcommand{\climaInsuranceMulti}{\textsc{Clima-Ins}\xspace}
\newcommand{\climaText}{\textsc{ClimaText}\xspace}
\newcommand{\climateStance}{\textsc{ClimateStance}\xspace}
\newcommand{\climateEng}{\textsc{ClimateEng}\xspace}
\newcommand{\climateFEVER}{\textsc{ClimateFEVER}\xspace}
\newcommand{\sciDCC}{\textsc{SciDCC}\xspace}
\newcommand{\expert}{climate change researcher\xspace}

\newcommand{\veryRel}{very relevant\xspace}

\newcommand{\cdp}{\textsc{CDP}\xspace}
\newcommand{\cdpCities}{\textsc{CDP-Cities}\xspace}
\newcommand{\cdpStates}{\textsc{CDP-States}\xspace}
\newcommand{\cdpCorps}{\textsc{CDP-Corp}\xspace}

\newcommand{\CLS}{$[CLS]$\xspace}

\newcommand{\clima}{\textsc{Clima-}\xspace}
\newcommand{\climate}{\textsc{Climate-}\xspace}

\title{Towards Answering Climate Questionnaires \\ from Unstructured Climate Reports}
\author{
  \begin{tabular}[t]{c@{\extracolsep{2em}}c}
    \textbf{Daniel Spokoyny}\textsuperscript{\textdagger}& \textbf{Tanmay Laud\textsuperscript{\ddag}} \\
  \end{tabular}
  \\
  \begin{tabular}[t]{c@{\extracolsep{2em}}c}
    \textbf{Thomas W. Corringham\textsuperscript{\ddag}} & \textbf{Taylor Berg-Kirkpatrick\textsuperscript{\ddag}} \\
  \end{tabular}
  \\
  \textsuperscript{\textdagger}Carnegie Mellon University
  \textsuperscript{\ddag}UC San Diego
}

\begin{document}
\maketitle
\begin{abstract}
The topic of Climate Change (CC) has received limited attention in NLP despite its urgency.
Activists and policymakers need NLP tools to effectively process the vast and rapidly growing unstructured textual climate reports into structured form.
To tackle this challenge we introduce two new large-scale climate questionnaire datasets and use their existing structure to train self-supervised models.
We conduct experiments to show that these models can learn to generalize to climate disclosures of different organizations types than seen during training.
We then use these models to help align texts from unstructured climate documents to the semi-structured questionnaires in a human pilot study.
Finally, to support further NLP research in the climate domain we introduce a benchmark of existing climate text classification datasets to better evaluate and compare existing models.\footnote{Corresponding Author:\texttt{dspokoyn@cs.cmu.edu}}
\end{abstract}
\section{Introduction}
\begin{SCfigure*}[][h]
  \centering
  \includegraphics[width=.7\textwidth]{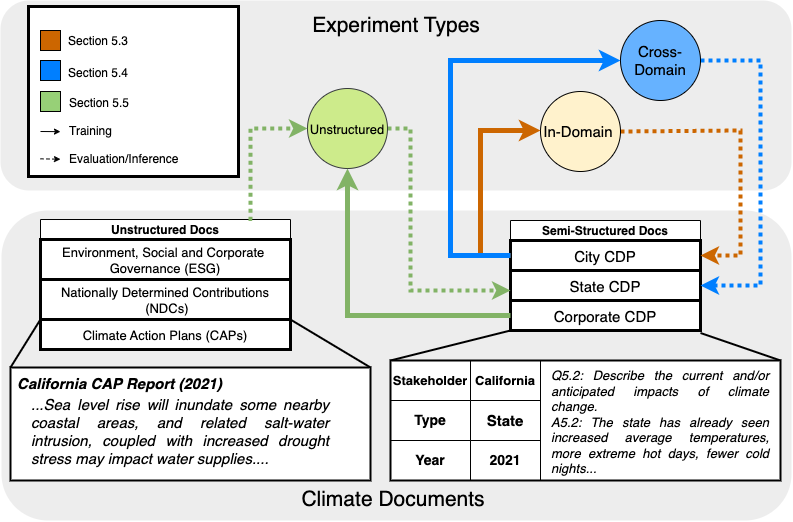}
  \caption{We conduct 3 experiments on \climaCDPQAClassification. In-Domain (\ref{exp2}) refers to training and evaluating on the same \nameTypeStakeholder. Cross-Domain (\ref{exp3}) refers to training and testing on different {\nameTypeStakeholder}s. 
  Finally, Unstructured Questionnaire Filling (\ref{exp4}) involves training on the whole \climaCDPQAClassification corpus and then using the model for mapping text from a \CAP report to a \cdp.
  We use solid and dashed arrows to denote training and inference/evaluation respectively.}
  \label{fig:tasks2}
\end{SCfigure*}

Globally, tens of thousands of climate reports have been generated by different \textit{stakeholders} such as corporations, cities, states, and national governments either voluntarily or in response to regulatory pressure.
These reports disclose vital information on carbon emissions, impacts, and risks -- for example, a firm's emissions reduction targets or a city's water risk and exposure to drought.
Increasingly, NLP is a critical technology supporting large scale processing of climate reports to enable downstream applications like detecting corporate greenwashing~\cite{Bingler2021CheapTA} or identifying misinformation about climate change~\cite{Meddeb2022CounteractingFF}.
However, for climate researchers to make use of the information contained in these \textit{unstructured} text documents, their contents must first be collated into \textit{semi-structured} questionnaires that have consistent fields across reporting bodies and report types.
These structured questionnaires, in turn, allow climate researchers to compare progress across different stakeholders and identify which areas need financing, education, policy changes or other resources.
Currently, this extraction process requires an immense amount of manual effort resulting in whole organizations focused on mapping a single type of unstructured reports (Nationally Determined Contribution) to a single type of semi-structured questionnaires (Sustainable Development Goals).\footnote{World Resources Institute's: www.climatewatchdata.org}\footnote{For more background info see Appendix~\ref{app:more-info}.}

In order to facilitate NLP research for this task, we introduce two new datasets, \climaCDP and \climaInsurance, which are composed of publicly accessible semi-structured questionnaires from different stakeholders including cities, states and corporations.
We utilize the structure of the questionnaires to train self-supervised classification models to align answers to questions (Section~\ref{exp2}).
Further, we show how the setup of our objective allows our model to generalize to a more challenging scenario where the set of questions and the stakeholder-type are both different at test time (Section~\ref{exp3}).
Finally, we show that models trained on \climaCDP can be directly applied to map passages from unstructured documents into questionnaire categories, which matches the real-world use-case that climate researchers need solved (Section~\ref{exp4}).
In Figure~\ref{fig:tasks2} we depict all three of these experiments as well as examples of the different reports and stakeholder-types.

There are other existing climate-specific datasets for detecting relevance to climate \cite{climatext}, identifying stance detection \cite{Vaid2022TowardsFC} and fact-checking \cite{climatefever} of social media claims.
In contrast, the questionnaires we introduce have an order of magnitude more data, are comprehensive in both the breadth of topics covered and the depth of detail provided making our models most suitable for a wide range of climate applications.

Climate reports have also been used as a source of unlabeled data to continue pretraining large language models to better adapt them for climate specific tasks \cite{Luccioni2020AnalyzingSR, climatebert}.
However, it remains an open question whether these domain-specific models can effectively generalize since evaluation of these models has been limited on the climate domain.
To address this gap in comprehensive evaluation, we collate five existing climate datasets, along with our two new datasets into a benchmark dataset (\climabench), and find that the domain-specific models like ClimateBERT underperform compared to existing general models (Section~\ref{exp1}).

In summary, our contributions are as follows:
\begin{enumerate}
\item We introduce two new datasets, \climaCDP and \climaInsurance, consisting of difficult classification tasks that are analogous to current manual work done by climate researchers, and conduct extensive in-domain experiments.
\item We collate and release \climabench, an evaluation dataset of climate-related text classification tasks and show that, counter-intuitively, general-purpose ML models outperform domain-specific models across tasks within the benchmark.
\item We conduct a pilot study, evaluated manually by a climate researcher, that uses a model trained on \climaCDP to populate a questionnaire from real-world unstructured climate reports.
\end{enumerate}

We believe our contributions are an important step for an emerging domain of building NLP tools for climate researchers.
To that end, we release our benchmark\footnote{\url{https://github.com/climabench/climabench}} and open-source our trained models\footnote{https://huggingface.co/climabench/miniLM-cdp-all} to encourage researchers to extend our existing datasets and contribute new ones.

\section{Related Work}
Climate policy evaluation is an active area of research in climate sciences where the goal is to evaluate the effectiveness of current climate policies so as to inform future policy decisions~\cite{Swarnakar2021NLPFC,cacao2021deeppolicytracker}. It allows for the development, assessment, and improvement of regulation, increases transparency and public support, and encourages public and private sector entities to make pledges or increase their levels of action~\cite{Fujiwara2019ThePO, ccml}. NLP has the potential to derive understandable insights from policy texts for these applications. 

Academic literature provides a valuable source of information for conducting these evaluation studies. However, a necessary first step is systematic evidence mapping or identifying which papers are relevant to a particular policy. \citet{BerrangFord2021SystematicMO}, for instance, build a machine learning system to filter scientific literature relevant to climate adaptation.

Another area of research involves utilizing unstructured climate documents for topic classification. \citet{corringham2021bert} attempt to use document headers from unstructured Nationally Determined Contribution (NDC) reports as coarse-grained labels to train a supervised classifier.
Most similar to our \climaQA work is \citet{Luccioni2020AnalyzingSR} who trained a model to map text passages from public financial disclosures to the 14 questions in Task Force on Climate-related Financial Disclosures (TCFD).
They recruited experts to manually label the text passages to the TCFD questions and only train their models on this labeled data.
Our work focuses on using the existing structure of large-scale public questionnaires to first train models and then apply them to unstructured texts.

NLP is also used to analyze social media data to understand public opinions and discourse around climate change~\citep{cc_micro}. 
\climaText~\cite{climatext} and \climateFEVER ~\cite{climatefever} extracted and filtered documents from Wikipedia and other sources to curate a CC corpus that was further annotated by humans.
In climate finance, \citet{Klbel2020DoesTC} have built NLP classifiers to distinguish texts describing physical climate risk versus transition risk.
While these studies have independently analyzed small annotated datasets, we make use of semi-structured disclosure forms comprising a much larger set of supervised data, made available to the CC and NLP communities in a clean and accessible format. Similar work has been conducted manually in the CC policy evaluation community (e.g., ClimateWatch) but not over the breadth and scope of documents we consider.

Finally, benchmarks have been an effective way to track progress and highlight the shortcomings of NLP models
in both general-purpose understanding (GLUE~\cite{glue}, SuperGLUE~\cite{superglue}) as well as specific domains such as legal NLP (LexGLUE~\cite{lexglue}) or biomedical NLP (BLURB~\cite{Gu2022DomainSpecificLM}).
\climabench follows on this chain of thought to provide a unified way to evaluate models on CC-specific problems.

\section{Datasets}%
\begin{table*}[t]
\centering
\resizebox{2\columnwidth}{!}{%
\begin{tabular}{p{2cm}p{15cm}p{5cm}c}
\toprule
 &
  \textbf{Free-form Text/Answer} &
  \textbf{Class / Question} & \textbf{\# Classes} \\ \toprule
\climaInsuranceMulti &
  ...Each year Aflac reports its US operations Scope 1 and Scope 2 emissions to the Carbon Disclosure Project. Since 2007, Aflac's owned facilities in terms of square feet have increased by more than 10\% while total Scope 1 and 2 CO2e emissions have significantly decreased compared to 2007 emissions...&
  Does the company have a plan to assess, reduce or mitigate its emissions in its operations or organizations? & 8\\ \hline
\climaCDPTopicClassification &
  ...These Plans must include management of CD\&E waste, both through on-site recycling and re-use and on-site waste processing prior to disposal.Westminster will contribute to the London Plan target of net self-sufficiency (managing 100\% of London’s waste within London) by 2026 by planning for Westminster’s apportionment targets... &
  Governance and Data Management & 12 \\ \hline
\climaCDPQAClassification (Cities) &
  Flooding from sea level rise will damage building and roads in the coastal neighborhoods of the city. Flooding also represents a risk to major transportation hubs infrastructure in the region. Coastal flooding can have a long-term effect on major industrial and commercial activities along the coastal areas of the city as well as damage urban forestry and local natural biodiversity. &
  Please describe the impacts experienced so far, and how you expect the hazard to impact in the future. & 294 \\ 
\bottomrule
\end{tabular}
}
\caption{\label{tab:examples}Examples (pairs of inputs and outputs) for the newly introduced datasets.}
\end{table*}

\begin{table*}[!ht]
\setlength\extrarowheight{10pt}
\centering
\resizebox{2\columnwidth}{!}{%
\begin{tabular}{lllllrrrr}
\hline
\textbf{Dataset} &
  \textbf{Source} &
  \textbf{Task Type} &
  \textbf{Domain} &
  \textbf{Stakeholder} &
  \textbf{\# Train} &
  \textbf{\# Dev} &
  \textbf{\# Test} &
  \textbf{\# Classes} \\ \hline
\climaInsuranceMulti & Ours                  & Multi-class Classification & NAIC            & Corporations          & 13.7K & 1.7K & 1.7K & 8  \\ \hline
 \climaCDPTopicClassification        & Ours                  & Topic Classification       & CDP         & Cities     & 46.8K & 8.7K & 8.9K & 12 \\ \hline
                      &                       &                            &                 & Cities     & 48.2K & 8.5K & 9.3K & 294  \\ \cline{5-9} 
 \climaCDPQAClassification &
 Ours &
Question Answering &
CDP &
  States &
 8.7K &
 0.9K &
1.1 K &
132 \\ \cline{5-9} 
 &
 &
 &
 &
  Corporations &
 34.5K &
 3.6K &
 4.9K &
 43 \\ \hline
 \climaText             & \citet{climatext}     & Binary Classification      & Wikipedia, 10-K & -          & 6K    & 0.3K & 1.6K & 2  \\ \hline
 \climateStance         & \citet{climatestance} & Ternary Classification     & Twitter         & -          & 3.0 K & 0.3K & 0.3K & 3  \\ \hline
 \climateEng            & \citet{climatestance} & Multi-class Classification & Twitter         & -          & 3K    & 0.3K & 0.3K & 5  \\ \hline
 \climateFEVER          & \citet{climatefever}  & Fact-Checking              & Wikipedia       & -          & -     & -    & 1.5K & 3  \\ \hline
 \sciDCC                & \citet{scidcc}        & Topic Classification       & Science Daily   & -          & 9.2K  & 1.1K & 1.1K & 20 \\ \hline
\end{tabular}%
}
\caption{General statistics of the datasets collected for \climabench and \climaCDPQAClassification.}
\label{tab:statistics}
\end{table*}

In this section we first describe our two new questionnaire datasets, \climaCDP and \climaInsurance, and then present all the text classification datasets we collected into \climabench.
We consider a questionnaire, a \nameSemiDoc, filled out by a \nameStakeholder for a particular year to have a set of questions and answers, ($\varQuestions$, $\varAnswers$) where the i-th question-answer pair \{$\varQuestion_i$, $\varAnswer_i$\} are both free-form text. 
Table~\ref{tab:examples} lists a few interesting examples from the newly introduced datasets.
The overall statistics of each dataset are given in Table~\ref{tab:statistics}, the token length distribution is given in Appendix Table~\ref{tab:data_token_stats} and details are explained below.

\subsection{\climaInsurance}
The annual NAIC Climate Risk Disclosure Survey\footnote{https://interactive.web.insurance.ca.gov} is a U.S. insurance regulation tool where insurers file non-confidential disclosures of their assessments and management of climate-related risks.
The purpose of the survey is to enhance transparency about how insurers manage climate-related risks and opportunities to enable better-informed collaboration on climate-related issues.
The dataset contains survey responses for the years 2012-2021, where each survey consists of eight questions all shown in Appendix~\ref{sec:appendix} and examples in Table~\ref{tab:examples}.
Companies have an option to fill the survey individually or as a group (in case of a conglomerate).
In the case of group filing, there may be duplicate answers repeated across all subsidiaries. We remove such responses resulting in a total of 17K question-answer pairs. Further, we delete the first sentence in each response as it contains obvious markers (like "Yes, we do X." or "No, we do not participate in Y."). The splits for training, validation and testing (80\%, 10\%, 10\%) are created by stratifying based on the company so that similar responses from the same company are not seen during train and test.

\subsection{\climaCDP}
Carbon Disclosure Project (CDP) is an international organisation that runs a global disclosure questionnaire for various stakeholders to report their environmental information.
In 2021 alone over 14,000 organizations filled out the questionnaire which contains hundreds of unique questions.

The \climaCDP, $\varCDPDataset$, is composed of parts [$\varCDPCities$, $\varCDPCompanies$, $\varCDPStates$] where each part is a set of questionnaires filled out by a city, company, or state respectively.
From the questionnaires we construct two tasks: topic classification (\climaCDPTopicClassification) and question classification (\climaCDPQAClassification).

\textbf{\climaCDPTopicClassification}
The CDP questionnaire contains a hierarchy of questions organized by topics such as \textit{energy}, \textit{food}, \textit{waste}.
We utilize these topics as \textit{labels} for a classification task and show the mapping in Appendix Table~\ref{tab:cdp_mapping}.
Thus, for each question-answer pair \{$\varQuestion_i$, $\varAnswer_i$\} we also have a topic label.
We formulate a topic classification task where the goal is to predict the \textit{topic} given the text of the \textit{answer}.

\textbf{\climaCDPQAClassification}
Our aim is to construct controlled experiments with proper evaluation metrics which closely resemble the real-world scenario of aligning unstructured climate reports to semi-structured ones.
For example, the $\nameCDPDataset$ allows us to test whether models can generalize to questionnaires of different \nameTypeStakeholder.
However, since the set of questions for each stakeholder type ($\varQuestions_{city}$, $\varQuestions_{corp}$, $\varQuestions_{state}$) are different from one another, a classifier predicting the question type will not be able to transfer to a new stakeholder type.
By using the text of the questions directly we can handle new questions at test time, which allows us to train on questionnaires from cities and test their generalization on questionnaires for states.
Since organization may file yearly reports which contain similar information we build train, dev and test splits stratified by the organizations.
Further we filter out duplicate, non-English, and short (less than 10 words) responses.

\subsection{\climabench}
\noindent In this section we introduce \climabench, a benchmark of climate related text classification tasks for evaluating NLP models. We collate five existing climate change related text datasets, described in detail below along with \climaInsurance and \climaCDPTopicClassification.

\textbf{\climaText} is a dataset for sentence-based climate change topic detection \citep{climatext}.
Each sentence is labelled indicating whether it is relevant to climate change or not.
Sentences were collected from the general web and Wikipedia as well as the climate-related risks section of US public companies' 10-K reports.

\textbf{\climateStance} and \textbf{\climateEng} \citet{climatestance} extracted Twitter data consisting of 3777 tweets posted during the 2019 United Nations Framework Convention on Climate Change.
Each tweet was labelled for two tasks: stance detection and categorical classification.
For the stance detection the authors labelled each tweet as \textit{In Favour}, \textit{Against} or \textit{Ambiguous} towards climate change prevention.
For categorical classification, the five classes are \textit{Disaster}, \textit{Ocean/Water}, \textit{Agriculture/Forestry}, \textit{Politics}, and \textit{General}.

\textbf{\climateFEVER}~\cite{climatefever} adopts the FEVER~\cite{Thorne18Fever} format for a fact-verification task based on climate change claims found on the Internet.
The dataset consists of 1,535 claims and five relevant evidence passages from Wikipedia for each claim.
The label set for each claim-evidence pair is \textit{Supports}, \textit{Refutes}, or \textit{Not Enough Info} for a total 7675 labelled examples.
For \climateFEVER, we concatenate the texts of each claim-evidence pair as a single input to the model.

\textbf{ \sciDCC}~\cite{scidcc} 
The Science Daily Climate Change or \sciDCC dataset is curated by scraping news articles from the Science Daily website~\cite{scidcc}. 
It contains around 11k news articles with 20 labeled categories relevant to climate change such as \textit{Earthquakes}, \textit{Pollution} and \textit{Hurricanes}.
Each article comprises of a title, a summary, and a body which on average is much longer (500-600 words) than the other climate text datasets.
For \sciDCC, we concatenate the text fields (title, summary and body) and provide a train, validation and test split  (80\%, 10\%, 10\%) for this data, ensuring the distribution of categories in the splits matches the overall distribution.

\section{Models}
Next, we are going to describe the various baselines and models that we use to conduct experiments using the datasets described above.
Most tasks are classification tasks that require in-domain finetuning.
For the text classification tasks in \climabench, we examine Transformer-based~\cite{transformer} pre-trained language models like BERT~\cite{bert}, RoBERTa~\cite{roberta}, distilled versions like DistilRoBERTa~\cite{distilroberta}, longer context models like Longformer~\cite{longformer}, and domain specific models like ClimateBERT~\cite{climatebert} and SciBERT~\cite{scibert}.
This helps us contrast the effects of model architecture, input length and in-domain pretraining on downstream tasks.
We provide more details about models in Appendix Section~\ref{transformer_models} and Table~\ref{tab:models}.
For a baseline, we consider a linear kernel Support Vector Machine (SVM) trained using TF-IDF transformed n-gram ({1,2,3}-gram) features.
We also include a simple Majority and Random class voting baselines.

For experimentation on \climaCDPQAClassification we consider a pre-trained Cross-Encoder MiniLM~\cite{wang2020minilm} model which was separately finetuned on the MS MARCO Passage Retrieval Dataset~\cite{msmarco} by \citet{reimers}.
The MS MARCO dataset contains real user queries together with annotated relevant text passages.
The model takes in as input the query concatenated with the passage and is trained to predict the pair's binary relevance score.
This model achieved state of the art performance across many retrieval tasks~\cite{beir}.
We consider this as a strong general purpose model in contrast to ClimateBERT which is a domain specific model.

\section{Experiments}
In our work we conduct four experiments: \textbf{(1)} \nameExpOne, \textbf{(2)} \nameExpTwo, \textbf{(3)} \nameExpThree, and \textbf{(4)} \nameExpFour.
For the first experiment, we examine the performance of existing general models as well as climate-specific models on our new \climabench evaluation dataset.
Experiments 2 and 3 focus on how we can utilize the semi-structured \climaQA dataset to create a self-supervised version of the unstructured document alignment task in a controlled setting with proper evaluation metrics.
Finally, in experiment 4 we will evaluate using human relevance judgements a model trained using the semi-structured CDP dataset can aid in aligning an unstructured climate report to the CDP questionnaire.

\subsection{Task Learning Details}
Each task has its own supervised training data that allows for in-domain finetuning for the target classification task.
In all experiments for all transformer models except MiniLM, we will add a classification head and do full finetuning.
For all the pre-trained models, we use publicly available Hugging Face ~\cite{huggingface} checkpoints.\footnote{We use the *-base configuration of each pre-trained model, i.e., 12 Transformer blocks, 768 hidden units, and 12 attention heads. For ClimateBERT we report scores for the F variant model on Huggingface. For the QA Cross-encoder, we use the MiniLM (12 layer, 384 hidden-unit) finetuned on MSMARCO available at \url{https://huggingface.co/cross-encoder/ms-marco-MiniLM-L-12-v2}}
For the Longformer, we use the default settings (windows of 512 tokens and a single global \CLS token).
We use the Scikit-learn API~\cite{scikit-learn} for the simple classifiers (Random and Majority class) and TF-IDF-based linear SVM models.  We grid-search the hyper parameters for SVM with 5-fold validation (Table~\ref{tab:svm_params}).

We use a training batch size of 32 and optimize using AdamW~\cite{adamw} with a learning rate of 5e-5 (linear warm-up ratio of 0.1, weight decay of 0.01) for 10 epochs with early stopping based on performance on development data (F1).
We use mixed precision (fp16), gradient checkpointing and gradient accumulation steps of 2 to train models efficiently on the limited compute (Appendix~\ref{compute}).
We truncate the input text when it exceeds the maximum input length of the model and otherwise pad the input.

\subsection{Text Classification on \climabench}\label{exp1}%
\begin{SCtable*}[1]
\centering
\setlength\extrarowheight{7pt}
\resizebox{.8\textwidth}{!}{%
\begin{tabular}{llllllrrrr}
\toprule
\multicolumn{1}{l}{} &
  \multicolumn{1}{l}{\textbf{\clima}} &
  \multicolumn{1}{l}{\textbf{\textsc{Cdp}}} &
  \multicolumn{1}{l}{\textbf{\clima}} &
  \multicolumn{1}{l}{\textbf{\climate}} &
  \multicolumn{1}{l}{\textbf{\climate}} &
  \multicolumn{1}{l}{} &
  \multicolumn{1}{l}{\textbf{\climate}} &
   \\ 
  \multicolumn{1}{l}{\textbf{Models}} &
  \multicolumn{1}{l}
 {\textbf{\textsc{Ins}}} &
  \multicolumn{1}{l}{\textbf{\textsc{Topic}}} &
  \multicolumn{1}{l}{\textbf{\textsc{Text}}} &
  \multicolumn{1}{l}{\textbf{\textsc{Stance}}} &
  \multicolumn{1}{l}{\textbf{\textsc{Eng}}} &
  \multicolumn{1}{l}{\textbf{\sciDCC}} &
  \multicolumn{1}{l}{\textbf{\textsc{FEVER}}} &
  \multicolumn{1}{l}{\textbf{\textsc{Avg.}}}\\\toprule
\multicolumn{1}{l}{Majority} &
  \multicolumn{1}{l}{4.11} &
  \multicolumn{1}{l}{3.65} &
  \multicolumn{1}{l}{42.08} &
  \multicolumn{1}{l}{29.68} &
  \multicolumn{1}{l}{13.83} &
  \multicolumn{1}{l}{0.79} &
  \multicolumn{1}{l}{26.08} &
  \multicolumn{1}{l}{20.10} \\ \hline
\multicolumn{1}{l}{Random} &
  \multicolumn{1}{l}{12.14} &
  \multicolumn{1}{l}{6.45} &
  \multicolumn{1}{l}{46.86} &
  \multicolumn{1}{l}{25.52} &
  \multicolumn{1}{l}{16.71} &
  \multicolumn{1}{l}{5.05} &
  \multicolumn{1}{l}{30.62} &
  \multicolumn{1}{l}{24.09} \\ \hline
\multicolumn{1}{l}{SVM} &
  \multicolumn{1}{l}{\textbf{86.00}} &
  \multicolumn{1}{l}{58.34} &
  \multicolumn{1}{l}{83.39} &
  \multicolumn{1}{l}{42.92} &
  \multicolumn{1}{l}{51.81} &
  \multicolumn{1}{l}{48.02} &
  \multicolumn{1}{c}{-} &
  \multicolumn{1}{c}{-} \\ \toprule
\multicolumn{1}{l}{BERT} &
  \multicolumn{1}{l}{84.57} &
  \multicolumn{1}{l}{64.64{$^\dagger$}} &
  \multicolumn{1}{l}{87.04{$^\dagger$}} &
  \multicolumn{1}{l}{55.37{$^\dagger$}} &
  \multicolumn{1}{l}{71.78} &
  \multicolumn{1}{l}{54.74{$^\dagger$}} &
  \multicolumn{1}{l}{62.47{$^\dagger$}} &
  \multicolumn{1}{l}{70.57{$^\dagger$}} \\ \hline
\multicolumn{1}{l}{RoBERTa} &
  \multicolumn{1}{l}{85.61{$^\dagger$}} &
  \multicolumn{1}{l}{\textbf{65.22}} &
  \multicolumn{1}{l}{85.97} &
  \multicolumn{1}{l}{\textbf{59.69}} &
  \multicolumn{1}{l}{\textbf{74.58}} &
  \multicolumn{1}{l}{52.90} &
  \multicolumn{1}{l}{60.74} &
  \multicolumn{1}{l}{\textbf{71.14}} \\ \hline
\multicolumn{1}{l}{DistilRoBERTa} &
  \multicolumn{1}{l}{84.38} &
  \multicolumn{1}{l}{63.61} &
  \multicolumn{1}{l}{86.06} &
  \multicolumn{1}{l}{52.51} &
  \multicolumn{1}{l}{72.33{$^\dagger$}} &
  \multicolumn{1}{l}{51.13} &
  \multicolumn{1}{l}{61.54} &
  \multicolumn{1}{l}{69.27} \\ \hline
\multicolumn{1}{l}{Longformer} &
  \multicolumn{1}{l}{84.35} &
  \multicolumn{1}{l}{64.03} &
  \multicolumn{1}{l}{\textbf{87.80}} &
  \multicolumn{1}{l}{34.68} &
  \multicolumn{1}{l}{72.28} &
  \multicolumn{1}{l}{\textbf{54.79}} &
  \multicolumn{1}{l}{60.82} &
  \multicolumn{1}{l}{67.72} \\ \hline
\multicolumn{1}{l}{SciBERT} &
  \multicolumn{1}{l}{84.43} &
  \multicolumn{1}{l}{63.62} &
  \multicolumn{1}{l}{83.29} &
  \multicolumn{1}{l}{48.67} &
  \multicolumn{1}{l}{70.50} &
  \multicolumn{1}{l}{51.83} &
  \multicolumn{1}{l}{\textbf{62.68}} &
  \multicolumn{1}{l}{68.45} \\ \hline
\multicolumn{1}{l}{ClimateBERT} &
  \multicolumn{1}{l}{84.80} &
  \multicolumn{1}{l}{64.24} &
  \multicolumn{1}{l}{85.14} &
  \multicolumn{1}{l}{52.84} &
  \multicolumn{1}{l}{71.83} &
  \multicolumn{1}{l}{52.97} &
  \multicolumn{1}{l}{61.54} &
  \multicolumn{1}{l}{69.44} \\ \bottomrule
\end{tabular}
}
\caption{Macro F1 Scores on the Classification Datasets. \textbf{Bold} and $\dagger$ indicate first and second highest performing model respectively. RoBERTa scores the best on average followed by BERT and ClimateBERT.}
\label{tab:classification}
\end{SCtable*}
In this section we use the new text classification \climabench dataset as an evaluation framework to compare the performance of the different models.
We use macro-averaged F1 score as our evaluation metric since the datasets are imbalanced and all classes are equally important.
For the pre-trained transformer models, we add a single linear classification layer on top of the final \CLS token representation 
and use a weighted cross-entropy loss with class balanced weights.\footnote{We do not evaluate linear models on fact-checking or QA as the heterogeneity of the input in these tasks do not align with the linear setup.}

\subsubsection{Results on \climabench}
We report text classification results on \climabench in Table~\ref{tab:classification} as well as an average across all tasks.
We find there is no single model that does the best across the board, but RoBERTa is a clear winner as it beats the other baselines on four out of eight tasks.
Both of the domain adapted models, SciBERT and ClimateBERT do worse than their non-adapted counterparts.
For example, ClimateBERT and the model it was warm-started from, DistilRoBERTa, are very similar in performance.
Overall, the transformer models have significantly better gains over linear ones except on \climaInsuranceMulti where the TF-IDF+SVM model is superior.
It shows that simple word co-occurrence statistics are enough for certain tasks and deep language models might not be the right solution in such cases.

\subsection{In-Domain \climaCDPQAClassification}\label{exp2}
\begin{table}[!ht]
\centering
\resizebox{1\columnwidth}{!}{%
\begin{tabular}{@{}rrrrrrrl@{}}
\toprule
\textbf{} &
  \multicolumn{1}{|c|}{\textbf{\cdpCities}} &
  \multicolumn{1}{c|}{\textbf{\cdpStates}} &
  \multicolumn{1}{c}{\textbf{\cdpCorps}}
   \\ \midrule
\multicolumn{1}{r|}{\textbf{Model}} &
  \multicolumn{1}{r|}{\textbf{MRR@10}} &
  \multicolumn{1}{r|}{\textbf{MRR@10}} &
  \multicolumn{1}{r}{\textbf{MRR@10}}
   \\ \midrule
& \multicolumn{3}{c}{\textbf{No Finetuning on \cdp}} &
   \\ \midrule
\multicolumn{1}{r|}{\textbf{BM25}} &
  \multicolumn{1}{r|}{0.055} &
  \multicolumn{1}{r|}{0.084} &
  \multicolumn{1}{r}{0.153}
   \\ \midrule
\multicolumn{1}{r|}{\textbf{MiniLM}} &
  \multicolumn{1}{r|}{0.099} &
  \multicolumn{1}{r|}{0.120} &
  \multicolumn{1}{r}{0.320}
   \\ \midrule
& \multicolumn{3}{c}{\textbf{Finetuned on \cdp}} &
   \\ \midrule
\multicolumn{1}{r|}{\textbf{}} &
  \multicolumn{1}{r|}{In-Domain} &
  \multicolumn{1}{r|}{In-Domain} &
  \multicolumn{1}{r}{In-Domain}
   \\ \midrule
\multicolumn{1}{r|}{\textbf{ClimateBERT}} &
  \multicolumn{1}{r|}{0.331} &
  \multicolumn{1}{r|}{0.422} &
  \multicolumn{1}{r}{0.753}
   \\ \midrule
\multicolumn{1}{r|}{\textbf{MiniLM}} &
  \multicolumn{1}{r|}{\textbf{0.366}} &
  \multicolumn{1}{r|}{0.482} &
  \multicolumn{1}{r}{\textbf{0.755}}
   \\ \midrule
& \multicolumn{3}{c}{\textbf{Best Model Finetuned on all}} &
   \\ \midrule
\multicolumn{1}{r|}{\textbf{MiniLM}} &
  \multicolumn{1}{r|}{0.352} &
  \multicolumn{1}{r|}{\textbf{0.489}} &
  \multicolumn{1}{r}{0.745}
   \\ \bottomrule
\end{tabular}%
}
\caption{MRR@10 scores for BM25, ClimateBERT and MSMARCO-MiniLM on the three subsets of \climaQA. Models finetuned and evaluated on same subset fall under In-Domain.}
\label{tab:mrr_scores}
\end{table}

We utilize the semi-structured nature of the questionnaire to train models in self-supervised fashion.
Specifically, we concatenate the free-form text of the answer and question and train a binary classifier to predict whether, in fact, the input answer matches the input question -- i.e. does $\varAnswer_i$, the $i$th answer in our dataset, provide an answer to $\varQuestion_j$, the $j$th question in our dataset:
$p(y_{ij}=1|\varQuestion_j, \varAnswer_i)$.
Since we assume that the indices are setup so that $ \varAnswer_i$ matches $\varQuestion_j$ if and only if $i = j$, the ground truth labels are given by $y^{*}_{ij}=\mathbbm{1}[i=j]$.

We use the filled out questionnaires as positive or relevant pairs and randomly sample five negative QA pairs for each relevant pair.
We train separate models on each \nameTypeStakeholder partition of the $\nameCDPDataset$ and evaluate them on the corresponding in-domain test sets.
During inference time, given an answer we compute a relevance score for all combinations of QA pairs from the full set of questions of a particular \nameTypeStakeholder.
$$\textrm{argmax}_{j \in \{1, \ldots, |\varQuestions|\}}\ \ p(y_{ij} = 1 | q_j, a_i)$$
Since there is a large number of questions, instead of accuracy we consider the Mean Reciprocal Rank at $k$ (MRR@$k$) scores for the top $k$ items returned by a model.
MRR, a popular metric used in the Information Retrieval field, is the average of the reciprocal ranks of results for a sample of queries where the relevance grading is binary (Yes/No).

We narrow down to two models, MiniLM and the ClimateBERT model to study the effects of fine-tuning and transfer learning on the three subdomains: \cdpCities, \cdpStates and \cdpCorps.
We also use BM25~\cite{bm25} and MiniLM with no training as baselines.

\subsubsection{Results}
We report the results of our in-domain experiments on \climaQA in Table~\ref{tab:mrr_scores} (detailed results in Appendix Table~\ref{tab:mrr_scores_full}).
We find that MiniLM, a much smaller model, beats ClimateBERT across all three different subsets.
It is hard to diagnose the exact reason why domain adaptation does not help in this case as well since the data used to further pretrain ClimateBERT is non-public.
There may be further room for improvement in domain adaptation for the MiniLM, but we leave this as future work.
Lastly, the best performing model, MiniLM, when finetuned on all three subsets, achieves comparable performance on Cities and Corporations while ranking highest on States.

\subsection{Transfer \climaCDPQAClassification}\label{exp3}
\begin{table}[!ht]
\centering
\resizebox{.4\textwidth}{!}{%
\begin{tabular}{@{}rrrrrr@{}}
\toprule
\multicolumn{1}{r|}{\textbf{}} &
  \multicolumn{1}{c|}{\textbf{\cdpStates}} &
  \multicolumn{1}{c}{\textbf{\cdpCorps}} \\ \midrule
\multicolumn{1}{r|}{\textbf{Model}} &
  \multicolumn{1}{r|}{\textbf{MRR@10}} &
  \multicolumn{1}{r}{\textbf{MRR@10}} \\ \midrule
& \multicolumn{2}{c}{\textbf{No Finetuning}} \\ \midrule
\multicolumn{1}{r|}{\textbf{BM25}} &
  \multicolumn{1}{r|}{0.084} &
  \multicolumn{1}{r}{0.153}\\ \midrule
\multicolumn{1}{r|}{\textbf{MiniLM}} &
  \multicolumn{1}{r|}{0.120} &
  \multicolumn{1}{r}{0.320} \\ \midrule
& \multicolumn{2}{c}{\textbf{Finetuned on \cdpCities}} \\ \midrule
\multicolumn{1}{r|}{\textbf{}} &
  \multicolumn{1}{r|}{Transfer} &
  \multicolumn{1}{r}{Transfer} \\ \midrule
\multicolumn{1}{r|}{\textbf{ClimateBERT}} &
  \multicolumn{1}{r|}{0.298} &
  \multicolumn{1}{r}{0.465} \\ \midrule
\multicolumn{1}{r|}{\textbf{MiniLM}} &
  \multicolumn{1}{r|}{\textbf{0.353}} &
  \multicolumn{1}{r}{\textbf{0.489}} \\ \bottomrule
\end{tabular}%
}
\caption{MRR@10 scores for BM25, ClimateBERT and MiniLM on the Transfer experiments. Models are finetuned on \cdpCities and evaluated on States and Corporations.}
\label{tab:mrr_scores_transfer}
\end{table}
In this section we explore whether it is possible for transfer learning to adapt to questionnaire from a new unseen \nameTypeStakeholder.
Since the $\varCDPCities$ dataset is the largest we use this partition as our training data.
Furthermore, since we have the ground truth questionnaires for both states and corporations we are able to evaluate the performance in a controlled setting.
At test time we follow the same procedure as for the in-domain experiment however, we marginalize over the set of questions from the unseen \nameTypeStakeholder.

\subsubsection{Results}
We summarize the MRR@$k$ ($k$=10) results for the transfer learning experiments in Table~\ref{tab:mrr_scores_transfer} (detailed results in Appendix Table~\ref{tab:mrr_scores_transfer_full}).
We show that both models are able to beat the no-training baselines.
We again find that the MiniLM model outperforms the ClimateBERT model across both transfer learning scenarios.
We do observe a significant drop in performance as compared to the in-domain finetuning experiments.
This gap is the largest for the corporations dataset, where the MRR@10 drops from 0.745 to 0.48.
Overall, we find that the transfer learning models are able to adapt to the unseen \nameTypeStakeholder but that there is still room for improvement.

\begin{table*}[t!]
\resizebox{2\columnwidth}{!}{
\begin{tabular}{lllp{11cm}}
\toprule
Ex. &
   &
  \textbf{Text Segment from State Climate Action Plans} &
  \textbf{Top Questions from \cdpStates (using fine-tuned MiniLM)} \\ \midrule

\multirow{2}{*}{1} &
   &
  \multirow{2}{11cm}{Sea level rise will inundate some nearby coastal areas, and related salt-water intrusion, coupled with increased drought stress may impact water supplies.} &
  Q1: Please describe the current and/or anticipated impacts of climate change. \\
 &
   &
  &
  Q3: Please detail any compounding factors that may worsen the impacts of climate change in your region. \\ \midrule

\multirow{2}{*}{2} &
   &
  \multirow{2}{11cm}{The afforestation goal is to increase the area of forested lands in the state by 50,000 acres annually through 2025.} &
  Q1: Please provide the details of your region's target(s). \\
 &
   &
  &
  Q2: Please provide details of your climate actions in the Land use sector. \\ \midrule

\multirow{2}{*}{3} &
   &
  \multirow{2}{11cm}{State law defines environmental justice as the fair treatment of people of all races, cultures, and incomes with respect to the development, adoption, implementation, and enforcement of environmental laws, regulations, and policies.} &
  Q1: Please explain why you do not have policies on deforestation and/or forest degradation. \\
 &
   &
  &
  Q4: Please provide details of your climate actions in the Governance sector. \\ 

  \hline
\end{tabular}}
\caption{Examples from our human pilot study in which our climate expert has evaluated the relevance of CDP questions linked to selected text from state climate action plans. A fragment of the matched text is presented with two illustrative questions from the set of five question matches generated by our model.}\label{cherry}
\end{table*}

\subsection{Questionnaire Filling}\label{exp4}

\begin{table}[t]
\resizebox{1\columnwidth}{!}{
\begin{tabular}{@{}lrrrrr@{}}
\toprule
 & Prec@1  & Prec@2  & Prec@3 & Prec@4 & Prec@5  \\ \midrule
Relevant      & 63.0 & 67.0 & 68.6 & 69.5 & 71.0  \\
Highly Relevant & 30.0 & 32.0 & 32.3 & 32.5 & 35.6  \\ \bottomrule
\end{tabular}}
\caption{Precision@$K$: We report the fraction of items in the top $K$ ranked retrievals that are either marked as highly relevant, or at least relevant, averaged across text examples. Relevance judgements were performed manually by an expert annotator.}
\label{tab:human-eval-YYY}
\end{table}

In our final experiment we consider the task of filling in a questionnaire based on an \textit{unstructured} text document -- specifically, we assume a State's Climate Action Plan (CAP) is available but the corresponding structured \cdp report is not.
Typically the \claps are much longer ($\sim$100 pages) and more comprehensive than any particular disclosure report.
The \claps include quantitative data, such as emission values or renewable electricity generation capacity, and qualitative data such as specific policy interventions across different sectors.
Populating CDP questionnaires allows for consistent comparisons to existing datasets which could further be used to compare strategies, identify gaps, or rank jurisdictions on the content and level of ambition in their stated plans.
However, this process is time-consuming and requires expert manual effort.

We select our best model, MiniLM, finetuned on the full \climaCDP dataset to conduct our \nameExpFour.
We can consider a State \CAP as an unstructured document $D_{un}$, to be a collection of texts, $D_{un} = \{t_1, t_2, \ldots, t_n\}$, where $t_i$ is a text segment.
The task is then to align a text segment $t_i$ to its corresponding CDP-State question $q_j \in \varQuestions_{state}$, i.e.
$\textrm{argmax}_{j \in \{1, \ldots, |\varQuestions_{state}|\}}\ \ p(y_{ij} = 1 | q_j, a_i)$.
Since we do not have the ground truth alignment we use a \expert in a procedure as follows: 
1) First, the expert (climate policy researcher on our team and co-author) selected 5 pages at random from a collection of 20 State \claps and then selected a random paragraph from each page as a text segment $t_i$.
2) Then, using our model we calculated relevance scores for each text segment question pair $(t_i, q_j)$ and selected the top 5 scoring questions for each text segment.
3) We then presented each segment along with the five questions to the \expert and had them annotate the relevance for each pair on a three point scale: No Relevance, Relevant, Highly Relevant.\footnote{By construction, in our rating there may be multiple relevant questions found for each text segment.}

\subsubsection{Human Evaluation}
Table~\ref{tab:human-eval-YYY} shows the \expert 's evaluation metrics for our model. 
Overall, 71.0\% of the 500 questions retrieved were judged \textit{Relevant} and 35.6\% rated \textit{Highly Relevant}.
One pitfall of our model is that there were more \veryRel predictions ranked fifth than first.
One possible explanation for this is that the top retrieved questions were often more general while the questions that were ranked lower were more specific and easier to match (see Table~\ref{tab:qs-dif} in the Appendix).

We show some examples of text segments and the selected questions in Table~\ref{cherry} and more in the Appendix Table~\ref{tab:cherry2}.
The first two examples show high degrees of success.
In example 1, our model correctly identifies the state CAP text as impact-related and captures the specific discussion of compound risks.
However, example 3 appears to highlight a gap in the CDP questionnaire related to the topic of environmental justice, a result in itself of considerable interest.
Although our pilot study is quite limited, it shows both the promise and the challenges of aligning unstructured climate documents to semi-structured questionnaires.

\section{Conclusion}
In summary, we introduced two climate questionnaire datasets and illustrated how using their existing structure we can train self-supervised models for climate question answering tasks analogous to real-world challenges faced by climate researchers.
Finally we lay the groundwork for future work in this domain by introducing a collated benchmark of existing climate text classification datasets.

\section{Limitations}
One current limitation of our benchmark is that the datasets are English only, thus restricting evaluation to English trained models.
Although the $\nameCDPDataset$ has disclosures in other languages it represents a small portion of the reports. We plan to include relevant climate change datasets from the multilingual European Union Public Data Catalog\footnote{data.europa.edu} in the future, while encouraging contributions from the broader community.
Another limitation is that for our human evaluation pilot study we were able to only get results for a single model.
We wish to build a small labeled dataset where climate experts map State climate action plans to their corresponding CDP questions for evaluation purposes.
Doing such manual labeling is particularly difficult for CDP due to the large number of questions but this resource could then be used efficiently to evaluate multiple models and baselines.

We do not thoroughly investigate the efficiency-accuracy trade-offs of the Transformer models in this work. We provide the compute and training efficiency statistics in \ref{emission_stats} and Table~\ref{tab:runtime_stats} as only a step in this direction.
In this work we used the MiniLM model, a cross-encoder, for the \climaCDPQAClassification experiments. 
Although this model is much smaller, at test time it requires a forward pass for each question-answer pair, which is computationally expensive.
In future work it would be interesting to compare the cross-encoder to bi-encoders model architectures to better understand the accuracy vs. performance trade-off.
We encourage future work on \climabench to leverage models that are both performant and efficient.

Finally, there exists more types of carbon disclosures  (TCFD, SBTi) as well as publicly accessible corporate sustainability reports that we wish to include but require more time-consuming scraping and data preprocessing.

\bibliography{anthology,custom}

\begin{thebibliography}{45}
\expandafter\ifx\csname natexlab\endcsname\relax\def\natexlab#1{#1}\fi

\bibitem[{esg(2004)}]{esg}
 2004.
\newblock \href
  {https://www.unglobalcompact.org/docs/issues_doc/Financial_markets/who_cares_who_wins.pdf}
  {Who cares wins: Connecting the financial markets to a changing world?}
\newblock Technical report, United Nations, The Global Compact.

\bibitem[{tcf(2022)}]{tcfd}
 2022.
\newblock \href
  {https://assets.bbhub.io/company/sites/60/2022/10/2022-TCFD-Status-Report.pdf}
  {2022 status report}.
\newblock Technical report, TCFD.

\bibitem[{cli(2022)}]{climatewatch}
 2022.
\newblock \href {https://www.climatewatchdata.org} {Climate watch}.
\newblock Technical report, World Resources Institute, Washington, D.C.

\bibitem[{Beltagy et~al.(2019)Beltagy, Lo, and Cohan}]{scibert}
Iz~Beltagy, Kyle Lo, and Arman Cohan. 2019.
\newblock \href {https://doi.org/10.18653/v1/D19-1371} {{S}ci{BERT}: A
  pretrained language model for scientific text}.
\newblock In \emph{Proceedings of the 2019 Conference on Empirical Methods in
  Natural Language Processing and the 9th International Joint Conference on
  Natural Language Processing (EMNLP-IJCNLP)}, pages 3615--3620, Hong Kong,
  China. Association for Computational Linguistics.

\bibitem[{Beltagy et~al.(2020)Beltagy, Peters, and Cohan}]{longformer}
Iz~Beltagy, Matthew~E. Peters, and Arman Cohan. 2020.
\newblock Longformer: The long-document transformer.
\newblock \emph{ArXiv}, abs/2004.05150.

\bibitem[{Berrang‐Ford et~al.(2021)Berrang‐Ford, Sietsma, Callaghan, Minx,
  Scheelbeek, Haddaway, Haines, and Dangour}]{BerrangFord2021SystematicMO}
Lea Berrang‐Ford, Anne~J Sietsma, Max~W. Callaghan, Jan~C. Minx, Pauline
  F.~D. Scheelbeek, Neal~Robert Haddaway, Andy Haines, and Alan~D. Dangour.
  2021.
\newblock Systematic mapping of global research on climate and health: a
  machine learning review.
\newblock \emph{The Lancet. Planetary Health}, 5:e514 -- e525.

\bibitem[{Bills et~al.(2022)Bills, Mackay, Deng-Beck, Bush, Kutner, Carless,
  and Bachra}]{cdp}
Amy Bills, Beth Mackay, Chang Deng-Beck, George Bush, Maia Kutner, Rachel
  Carless, and Simeran Bachra. 2022.
\newblock Protecting people and the planet: Putting people at the heart of city
  climate action.
\newblock Technical report, CDP.

\bibitem[{Bingler et~al.(2021)Bingler, Kraus, and
  Leippold}]{Bingler2021CheapTA}
Julia~Anna Bingler, Mathias Kraus, and Markus Leippold. 2021.
\newblock Cheap talk and cherry-picking: What climatebert has to say on
  corporate climate risk disclosures.
\newblock \emph{Corporate Finance: Governance}.

\bibitem[{Brown et~al.(2007)Brown, de~Jong, and Lessidrenska}]{gri}
Halina~Szejnwald Brown, Martin de~Jong, and Teodorina Lessidrenska. 2007.
\newblock The rise of the global reporting initiative (gri) as a case of
  institutional entrepreneurship.
\newblock Working Paper~36, John F. Kennedy School of Government, Harvard
  University.

\bibitem[{Campos et~al.(2016)Campos, Nguyen, Rosenberg, Song, Gao, Tiwary,
  Majumder, Deng, and Mitra}]{msmarco}
Daniel~Fernando Campos, Tri Nguyen, Mir Rosenberg, Xia Song, Jianfeng Gao,
  Saurabh Tiwary, Rangan Majumder, Li~Deng, and Bhaskar Mitra. 2016.
\newblock Ms marco: A human generated machine reading comprehension dataset.
\newblock \emph{ArXiv}, abs/1611.09268.

\bibitem[{Carroll(2009)}]{csr}
Archie~B. Carroll. 2009.
\newblock A history of corporate social responsibility: Concepts and practices.
\newblock In Andrew Crane, Dirk Matten, Abagail McWilliams, Jeremy Moon, and
  Donald~S. Siegel, editors, \emph{The Oxford Handbook of Corporate Social
  Responsibility}. Oxford University Press, Oxford.

\bibitem[{Cação et~al.(2021)Cação, Reali~Costa, Unterstell, Yonaha, Stec,
  and Ishisaki}]{cacao2021deeppolicytracker}
Flávio~N Cação, Anna~Helena Reali~Costa, Natalie Unterstell, Liuca Yonaha,
  Taciana Stec, and Fábio Ishisaki. 2021.
\newblock \href {https://www.climatechange.ai/papers/icml2021/35}
  {Deeppolicytracker: Tracking changes in environmental policy in the brazilian
  federal official gazette with deep learning}.
\newblock In \emph{ICML 2021 Workshop on Tackling Climate Change with Machine
  Learning}.

\bibitem[{Chalkidis et~al.(2022)Chalkidis, Jana, Hartung, Bommarito,
  Androutsopoulos, Katz, and Aletras}]{lexglue}
Ilias Chalkidis, Abhik Jana, Dirk Hartung, Michael~James Bommarito, Ion
  Androutsopoulos, Daniel~Martin Katz, and Nikolaos Aletras. 2022.
\newblock Lexglue: A benchmark dataset for legal language understanding in
  english.
\newblock In \emph{ACL}.

\bibitem[{Corringham et~al.(2021)Corringham, Spokoyny, Xiao, Cha, Lemarchand,
  Syal, Olson, and Gershunov}]{corringham2021bert}
Tom Corringham, Daniel Spokoyny, Eric Xiao, Christopher Cha, Colin Lemarchand,
  Mandeep Syal, Ethan Olson, and Alexander Gershunov. 2021.
\newblock \href {https://www.climatechange.ai/papers/icml2021/45} {Bert
  classification of paris agreement climate action plans}.
\newblock In \emph{ICML 2021 Workshop on Tackling Climate Change with Machine
  Learning}.

\bibitem[{Devlin et~al.(2019)Devlin, Chang, Lee, and Toutanova}]{bert}
Jacob Devlin, Ming-Wei Chang, Kenton Lee, and Kristina Toutanova. 2019.
\newblock \href {https://doi.org/10.18653/v1/N19-1423} {{BERT}: Pre-training of
  deep bidirectional transformers for language understanding}.
\newblock In \emph{Proceedings of the 2019 Conference of the North {A}merican
  Chapter of the Association for Computational Linguistics: Human Language
  Technologies, Volume 1 (Long and Short Papers)}, pages 4171--4186,
  Minneapolis, Minnesota. Association for Computational Linguistics.

\bibitem[{Fujiwara et~al.(2019)Fujiwara, van Asselt, B{\"o}$\beta$ner, Voigt,
  Spyridaki, Flamos, Alberola, Williges, T{\"u}rk, and ten
  Donkelaar}]{Fujiwara2019ThePO}
Noriko. Fujiwara, Harro van Asselt, Stefan B{\"o}$\beta$ner, Sebastian Voigt,
  Niki-Artemis Spyridaki, Alexandros Flamos, Emilie Alberola, Keith Williges,
  Andreas T{\"u}rk, and Michael ten Donkelaar. 2019.
\newblock The practice of climate change policy evaluations in the european
  union and its member states: results from a meta-analysis.
\newblock \emph{Sustainable Earth}, 2:1--16.

\bibitem[{Gu et~al.(2022)Gu, Tinn, Cheng, Lucas, Usuyama, Liu, Naumann, Gao,
  and Poon}]{Gu2022DomainSpecificLM}
Yuxian Gu, Robert Tinn, Hao Cheng, Michael~R. Lucas, Naoto Usuyama, Xiaodong
  Liu, Tristan Naumann, Jianfeng Gao, and Hoifung Poon. 2022.
\newblock Domain-specific language model pretraining for biomedical natural
  language processing.
\newblock \emph{ACM Transactions on Computing for Healthcare (HEALTH)}, 3:1 --
  23.

\bibitem[{Kirilenko and Stepchenkova(2014)}]{cc_micro}
Andrei~P. Kirilenko and Svetlana~O. Stepchenkova. 2014.
\newblock \href
  {https://doi.org/https://doi.org/10.1016/j.gloenvcha.2014.02.008} {Public
  microblogging on climate change: One year of twitter worldwide}.
\newblock \emph{Global Environmental Change}, 26:171--182.

\bibitem[{K{\"o}lbel et~al.(2020)K{\"o}lbel, Leippold, Rillaerts, and
  Wang}]{Klbel2020DoesTC}
Julian~F. K{\"o}lbel, Markus Leippold, Jordy Rillaerts, and Qian Wang. 2020.
\newblock Does the cds market reflect regulatory climate risk disclosures.

\bibitem[{Lacoste et~al.(2019)Lacoste, Luccioni, Schmidt, and
  Dandres}]{lacoste2019quantifying}
Alexandre Lacoste, Alexandra Luccioni, Victor Schmidt, and Thomas Dandres.
  2019.
\newblock Quantifying the carbon emissions of machine learning.
\newblock \emph{arXiv preprint arXiv:1910.09700}.

\bibitem[{Leippold and Diggelmann(2020)}]{climatefever}
Markus Leippold and Thomas Diggelmann. 2020.
\newblock \href {https://www.climatechange.ai/papers/neurips2020/67}
  {Climate-fever: A dataset for verification of real-world climate claims}.
\newblock In \emph{NeurIPS 2020 Workshop on Tackling Climate Change with
  Machine Learning}.

\bibitem[{Leippold and Varini(2020)}]{climatext}
Markus Leippold and Francesco~Saverio Varini. 2020.
\newblock \href {https://www.climatechange.ai/papers/neurips2020/69}
  {Climatext: A dataset for climate change topic detection}.
\newblock In \emph{NeurIPS 2020 Workshop on Tackling Climate Change with
  Machine Learning}.

\bibitem[{Liu et~al.(2019)Liu, Ott, Goyal, Du, Joshi, Chen, Levy, Lewis,
  Zettlemoyer, and Stoyanov}]{roberta}
Yinhan Liu, Myle Ott, Naman Goyal, Jingfei Du, Mandar Joshi, Danqi Chen, Omer
  Levy, Mike Lewis, Luke Zettlemoyer, and Veselin Stoyanov. 2019.
\newblock Roberta: A robustly optimized bert pretraining approach.
\newblock \emph{ArXiv}, abs/1907.11692.

\bibitem[{Loshchilov and Hutter(2019)}]{adamw}
Ilya Loshchilov and Frank Hutter. 2019.
\newblock Decoupled weight decay regularization.
\newblock In \emph{ICLR}.

\bibitem[{Luccioni et~al.(2020)Luccioni, Baylor, and
  Duch{\^e}ne}]{Luccioni2020AnalyzingSR}
Alexandra~Sasha Luccioni, Emily Baylor, and Nicolas~Anton Duch{\^e}ne. 2020.
\newblock Analyzing sustainability reports using natural language processing.
\newblock \emph{ArXiv}, abs/2011.08073.

\bibitem[{Meddeb et~al.(2022)Meddeb, Ruseti, Dascalu, Terian, and
  Travadel}]{Meddeb2022CounteractingFF}
Paul Meddeb, Stefan Ruseti, Mihai Dascalu, Simina Terian, and S{\'e}bastien
  Travadel. 2022.
\newblock Counteracting french fake news on climate change using language
  models.
\newblock \emph{Sustainability}.

\bibitem[{Mishra and Mittal(2021)}]{scidcc}
Prakamya Mishra and Rohan Mittal. 2021.
\newblock \href {https://www.climatechange.ai/papers/icml2021/76} {Neuralnere:
  Neural named entity relationship extraction for end-to-end climate change
  knowledge graph construction}.
\newblock In \emph{ICML 2021 Workshop on Tackling Climate Change with Machine
  Learning}.

\bibitem[{Pedregosa et~al.(2011)Pedregosa, Varoquaux, Gramfort, Michel,
  Thirion, Grisel, Blondel, Prettenhofer, Weiss, Dubourg, Vanderplas, Passos,
  Cournapeau, Brucher, Perrot, and Duchesnay}]{scikit-learn}
F.~Pedregosa, G.~Varoquaux, A.~Gramfort, V.~Michel, B.~Thirion, O.~Grisel,
  M.~Blondel, P.~Prettenhofer, R.~Weiss, V.~Dubourg, J.~Vanderplas, A.~Passos,
  D.~Cournapeau, M.~Brucher, M.~Perrot, and E.~Duchesnay. 2011.
\newblock Scikit-learn: Machine learning in {P}ython.
\newblock \emph{Journal of Machine Learning Research}, 12:2825--2830.

\bibitem[{Reimers and Gurevych(2019)}]{reimers}
Nils Reimers and Iryna Gurevych. 2019.
\newblock \href {https://arxiv.org/abs/1908.10084} {Sentence-bert: Sentence
  embeddings using siamese bert-networks}.
\newblock In \emph{Proceedings of the 2019 Conference on Empirical Methods in
  Natural Language Processing}. Association for Computational Linguistics.

\bibitem[{Robertson and Zaragoza(2009)}]{bm25}
Stephen Robertson and Hugo Zaragoza. 2009.
\newblock \href {https://doi.org/10.1561/1500000019} {The probabilistic
  relevance framework: Bm25 and beyond}.
\newblock \emph{Found. Trends Inf. Retr.}, 3(4):333–389.

\bibitem[{Rolnick et~al.(2022)Rolnick, Donti, Kaack, Kochanski, Lacoste,
  Sankaran, Ross, Milojevic-Dupont, Jaques, Waldman-Brown, Luccioni, Maharaj,
  Sherwin, Mukkavilli, Kording, Gomes, Ng, Hassabis, Platt, Creutzig, Chayes,
  and Bengio}]{ccml}
David Rolnick, Priya~L. Donti, Lynn~H. Kaack, Kelly Kochanski, Alexandre
  Lacoste, Kris Sankaran, Andrew~Slavin Ross, Nikola Milojevic-Dupont, Natasha
  Jaques, Anna Waldman-Brown, Alexandra~Sasha Luccioni, Tegan Maharaj, Evan~D.
  Sherwin, S.~Karthik Mukkavilli, Konrad~P. Kording, Carla~P. Gomes, Andrew~Y.
  Ng, Demis Hassabis, John~C. Platt, Felix Creutzig, Jennifer Chayes, and
  Yoshua Bengio. 2022.
\newblock \href {https://doi.org/10.1145/3485128} {Tackling climate change with
  machine learning}.
\newblock \emph{ACM Comput. Surv.}, 55(2).

\bibitem[{Sachs(2012)}]{sdg}
Jeffrey~D. Sachs. 2012.
\newblock \href {https://doi.org/10.1016/S0140-6736(12)60685-0} {From
  millennium development goals to sustainable development goals}.
\newblock \emph{The Lancet}, 379(9832):2206--2211.

\bibitem[{Sanh et~al.(2019)Sanh, Debut, Chaumond, and Wolf}]{distilroberta}
Victor Sanh, Lysandre Debut, Julien Chaumond, and Thomas Wolf. 2019.
\newblock Distilbert, a distilled version of bert: smaller, faster, cheaper and
  lighter.
\newblock \emph{ArXiv}, abs/1910.01108.

\bibitem[{Stede and Patz(2021)}]{stede-patz-2021-climate}
Manfred Stede and Ronny Patz. 2021.
\newblock \href {https://doi.org/10.18653/v1/2021.nlp4posimpact-1.2} {The
  climate change debate and natural language processing}.
\newblock In \emph{Proceedings of the 1st Workshop on NLP for Positive Impact},
  pages 8--18, Online. Association for Computational Linguistics.

\bibitem[{Swarnakar and Modi(2021)}]{Swarnakar2021NLPFC}
Pradip Swarnakar and Ashutosh Modi. 2021.
\newblock Nlp for climate policy: Creating a knowledge platform for holistic
  and effective climate action.
\newblock \emph{ArXiv}, abs/2105.05621.

\bibitem[{Thakur et~al.(2021)Thakur, Reimers, Ruckl'e, Srivastava, and
  Gurevych}]{beir}
Nandan Thakur, Nils Reimers, Andreas Ruckl'e, Abhishek Srivastava, and Iryna
  Gurevych. 2021.
\newblock Beir: A heterogenous benchmark for zero-shot evaluation of
  information retrieval models.
\newblock \emph{ArXiv}, abs/2104.08663.

\bibitem[{Thorne et~al.(2018)Thorne, Vlachos, Christodoulopoulos, and
  Mittal}]{Thorne18Fever}
James Thorne, Andreas Vlachos, Christos Christodoulopoulos, and Arpit Mittal.
  2018.
\newblock {FEVER}: a large-scale dataset for fact extraction and
  {VERification}.
\newblock In \emph{NAACL-HLT}.

\bibitem[{Vaid et~al.(2022{\natexlab{a}})Vaid, Pant, and
  Shrivastava}]{Vaid2022TowardsFC}
Roopal Vaid, Kartikey Pant, and Manish Shrivastava. 2022{\natexlab{a}}.
\newblock Towards fine-grained classification of climate change related social
  media text.
\newblock In \emph{ACL}.

\bibitem[{Vaid et~al.(2022{\natexlab{b}})Vaid, Pant, and
  Shrivastava}]{climatestance}
Roopal Vaid, Kartikey Pant, and Manish Shrivastava. 2022{\natexlab{b}}.
\newblock \href {https://doi.org/10.18653/v1/2022.acl-srw.35} {Towards
  fine-grained classification of climate change related social media text}.
\newblock In \emph{Proceedings of the 60th Annual Meeting of the Association
  for Computational Linguistics: Student Research Workshop}, pages 434--443,
  Dublin, Ireland. Association for Computational Linguistics.

\bibitem[{Vaswani et~al.(2017)Vaswani, Shazeer, Parmar, Uszkoreit, Jones,
  Gomez, Kaiser, and Polosukhin}]{transformer}
Ashish Vaswani, Noam Shazeer, Niki Parmar, Jakob Uszkoreit, Llion Jones,
  Aidan~N Gomez, \L~ukasz Kaiser, and Illia Polosukhin. 2017.
\newblock \href
  {https://proceedings.neurips.cc/paper/2017/file/3f5ee243547dee91fbd053c1c4a845aa-Paper.pdf}
  {Attention is all you need}.
\newblock In \emph{Advances in Neural Information Processing Systems},
  volume~30. Curran Associates, Inc.

\bibitem[{Wang et~al.(2019)Wang, Pruksachatkun, Nangia, Singh, Michael, Hill,
  Levy, and Bowman}]{superglue}
Alex Wang, Yada Pruksachatkun, Nikita Nangia, Amanpreet Singh, Julian Michael,
  Felix Hill, Omer Levy, and Samuel Bowman. 2019.
\newblock \href
  {https://proceedings.neurips.cc/paper/2019/file/4496bf24afe7fab6f046bf4923da8de6-Paper.pdf}
  {Superglue: A stickier benchmark for general-purpose language understanding
  systems}.
\newblock In \emph{Advances in Neural Information Processing Systems},
  volume~32. Curran Associates, Inc.

\bibitem[{Wang et~al.(2018)Wang, Singh, Michael, Hill, Levy, and Bowman}]{glue}
Alex Wang, Amanpreet Singh, Julian Michael, Felix Hill, Omer Levy, and Samuel
  Bowman. 2018.
\newblock \href {https://doi.org/10.18653/v1/W18-5446} {{GLUE}: A multi-task
  benchmark and analysis platform for natural language understanding}.
\newblock In \emph{Proceedings of the 2018 {EMNLP} Workshop {B}lackbox{NLP}:
  Analyzing and Interpreting Neural Networks for {NLP}}, pages 353--355,
  Brussels, Belgium. Association for Computational Linguistics.

\bibitem[{Wang et~al.(2020)Wang, Wei, Dong, Bao, Yang, and
  Zhou}]{wang2020minilm}
Wenhui Wang, Furu Wei, Li~Dong, Hangbo Bao, Nan Yang, and Ming Zhou. 2020.
\newblock \href {http://arxiv.org/abs/2002.10957} {Minilm: Deep self-attention
  distillation for task-agnostic compression of pre-trained transformers}.

\bibitem[{Webersinke et~al.(2021)Webersinke, Kraus, Bingler, and
  Leippold}]{climatebert}
Nicolas Webersinke, Mathias Kraus, Julia~Anna Bingler, and Markus Leippold.
  2021.
\newblock Climatebert: A pretrained language model for climate-related text.
\newblock \emph{ArXiv}, abs/2110.12010.

\bibitem[{Wolf et~al.(2020)Wolf, Debut, Sanh, Chaumond, Delangue, Moi, Cistac,
  Rault, Louf, Funtowicz, Davison, Shleifer, von Platen, Ma, Jernite, Plu, Xu,
  Scao, Gugger, Drame, Lhoest, and Rush}]{huggingface}
Thomas Wolf, Lysandre Debut, Victor Sanh, Julien Chaumond, Clement Delangue,
  Anthony Moi, Pierric Cistac, Tim Rault, Rémi Louf, Morgan Funtowicz, Joe
  Davison, Sam Shleifer, Patrick von Platen, Clara Ma, Yacine Jernite, Julien
  Plu, Canwen Xu, Teven~Le Scao, Sylvain Gugger, Mariama Drame, Quentin Lhoest,
  and Alexander~M. Rush. 2020.
\newblock \href {https://www.aclweb.org/anthology/2020.emnlp-demos.6}
  {Transformers: State-of-the-art natural language processing}.
\newblock In \emph{Proceedings of the 2020 Conference on Empirical Methods in
  Natural Language Processing: System Demonstrations}, pages 38--45, Online.
  Association for Computational Linguistics.

\end{thebibliography}
\bibliographystyle{acl_natbib}

\appendix
\section{Appendix}
\label{sec:appendix}
\subsection{Compute Details}
\label{compute}
We used a 24 core AMD Ryzen CPU machine with 128 GB RAM for data processing. For training and inference of the deep learning models, we utilize 4 Nvidia RTX 2080Ti GPUs with 11GB memory each. Each model was trained on a single GPU at a time.
\begin{table*}[ht]
\centering
\resizebox{1.5\columnwidth}{!}{%
\begin{tabular}{ll}
\hline
\textbf{Section}                               & \textbf{Category/Label}        \\ \hline
Hazards: Adaptation                            & Adaptation                     \\ \hline
Adaptation                                     & Adaptation                     \\ \hline
Buildings                                      & Buildings                      \\ \hline
Hazards: Climate Hazards                       & Climate Hazards                \\ \hline
Hazards: Social Risks                          & Climate Hazards                \\ \hline
Climate Hazards                                & Climate Hazards                \\ \hline
Climate Hazards and Vulnerability              & Climate Hazards                \\ \hline
Climate Hazards \& Vulnerability               & Climate Hazards                \\ \hline
City-wide Emissions                            & Emissions                      \\ \hline
Emissions Reduction                            & Emissions                      \\ \hline
GHG Emissions Data                             & Emissions                      \\ \hline
Local Government Emissions                     & Emissions                      \\ \hline
Emissions Reduction: City-wide                 & Emissions                      \\ \hline
City Wide Emissions                            & Emissions                      \\ \hline
Emissions Reduction: Local Government          & Emissions                      \\ \hline
Local Government Operations GHG Emissions Data & Emissions                      \\ \hline
Energy Data                                    & Energy                         \\ \hline
Energy                                         & Energy                         \\ \hline
Food                                           & Food                           \\ \hline
Governance and Data Management                 & Governance and Data Management \\ \hline
Opportunities                                  & Opportunities                  \\ \hline
Strategy                                       & Strategy                       \\ \hline
Urban Planning                                 & Strategy                       \\ \hline
Transport                                      & Transport                      \\ \hline
Waste                                          & Waste                          \\ \hline
Water                                          & Water                          \\ \hline
Water Security                                 & Water                          \\ \hline
\end{tabular}%
}
\caption{The section topics in the CDP Cities Questionnaire and the corresponding Labels assigned by a climate expert.}
\label{tab:cdp_mapping}
\end{table*}

\begin{table}[ht]
\centering
\resizebox{\columnwidth}{!}{%
\begin{tabular}{lrrrr}
\toprule
\textbf{Task}         & \textbf{Average} & \textbf{Max}  & \textbf{Min} & \textbf{Std} \\ \hline
\climaInsurance      & 203              & 4588 & 11           & 326          \\ \hline
\climaInsuranceMulti & 206              & 4588 & 11           & 335          \\ \hline
\climaCDP             & 73               & 801  & 11           & 83           \\ \hline
\climaQA         & 105              & 834  & 15           & 88           \\ \hline
\climaText             & 23               & 124  & 11           & 10           \\ \hline
\climateStance         & 30               & 98   & 11           & 12           \\ \hline
\climateEng            & 30               & 98   & 11           & 12           \\ \hline
\climateFEVER          & 47               & 311  & 11           & 19           \\ \hline
\sciDCC                & 580              & 2014 & 13           & 223          \\ \bottomrule
\end{tabular}%
}
\caption{Statistics for the number of tokens in each task of \climabench}
\label{tab:data_token_stats}
\end{table}

\begin{table*}[ht]
\centering
\begin{tabular}{@{}llrrr@{}}
\toprule
\textbf{Model} & \textbf{Source} & \textbf{\# Params} & \textbf{Vocab Size} & \textbf{Max Length} \\ \toprule
BERT          & \cite{bert} & 110M & 30K & 512  \\ \midrule
RoBERTa       & \cite{roberta} & 125M & 50K & 512  \\ \midrule
DistilRoBERTa & \cite{distilroberta}  & 82M  & 50K & 512  \\ \midrule
Longformer    & \cite{longformer} & 149M & 50K & 4096 \\ \midrule
SciBERT       & \cite{scibert} & 110M & 30K & 512  \\ \midrule
ClimateBERT   & \cite{climatebert} & 82M  & 50K & 512  \\ \bottomrule
\end{tabular}
\caption{Pretrained Transformer Language Models used for Classification tasks}
\label{tab:models}
\end{table*}
\subsection{CO2 Emission Related to Experiments}
\label{emission_stats}
A cumulative of 338 hours of computation was performed on hardware of type RTX 2080 Ti (TDP of 250W).
Total emissions are estimated to be 36.5 kgCO$_2$eq.
Estimations were conducted using the \href{https://mlco2.github.io/impact#compute}{MachineLearning Impact calculator} presented in \cite{lacoste2019quantifying}.

\begin{table*}[!ht]
\centering
\resizebox{2\columnwidth}{!}{%
\begin{tabular}{@{}lrrr@{}}
\toprule
\textbf{Model} &
  \multicolumn{1}{l}{\textbf{Avg. Runtime (in hours)}} &
  \multicolumn{1}{l}{\textbf{Avg. Train Samples/Second}} &
  \multicolumn{1}{l}{\textbf{Avg. Train Steps/Second}} \\ \toprule
ClimateBERT & 0.40  & 104.83 & 1.64 \\ \midrule
DistilRoBERTa                       & 0.40  & 101.04 & 1.58 \\\midrule
SciBERT                          & 0.70  & 53.86  & 0.84 \\\midrule
RoBERTa                             & 0.80  & 50.46  & 0.79 \\\midrule
BERT                    & 0.85  & 49.32  & 0.77 \\\midrule
Longformer             & 14.95 & 13.82  & 0.76 \\ \bottomrule
\end{tabular}%
}
\caption{Compute Efficiency Metrics for the Pretrained Transformer models for the experiments conducted on \climabench. Models based on the DistilRoBERTa architecture are the most efficient due to smaller model size.}
\label{tab:runtime_stats}
\end{table*}

\subsection{Pretrained Transformer Models}
\label{transformer_models}
\noindent\textbf{BERT}~\cite{bert} is a popular Transformer-based language model pre-trained on masked language modeling and next sentence prediction tasks. It makes use of WordPiece tokenization algorithm that breaks a word into several subwords, such that commonly seen subwords can also be represented by the model.

\noindent\textbf{RoBERTa}~\cite{roberta} uses dynamic masking and eliminates the next sentence prediction pre-training task, while using a larger vocabulary and pre-training on much larger corpora compared to BERT. Another notable difference is the use of byte pair encoding compared to wordPiece in BERT. 

\noindent\textbf{DistilRoBERTa}~\cite{distilroberta} leverages knowledge distillation during the pre-training phase reducing the size of the RoBERTa model by 40\%, while retaining 97\% of its language understanding capabilities and being 60\% faster. \citet{distilroberta} originally distilled the BERT model but we utilize the better performing RoBERTa version in our experiments.

\noindent\textbf{Longformer}~\cite{longformer} extends
Transformer-based models to support longer sequences with the help of sparse-attention. It uses a combination of local attention and global attention mechanism that allows for linear attention complexity and thus makes it feasible to run on longer documents (max 4096 tokens). It however takes much longer to train than the shorter context (512 tokens) models.

\noindent\textbf{SciBERT}~\cite{scibert}, a pretrained language model based on BERT, leverages unsupervised pretraining on a large multi-domain corpus of scientific publications to improve performance on downstream scientific NLP tasks. It was evaluated on tasks like sequence tagging, sentence classification and dependency parsing with datasets from scientific domains. SciBERT gives significant improvements over BERT on these datasets.

\noindent\textbf{ClimateBERT}~\cite{climatebert} was warm-started from the DistilRoBERTa model and pretrained on text corpora from climate-related research paper abstracts, corporate and general news and reports from companies that were not publicly released with the model. It was evaluated on tasks like sentiment analysis (using a private dataset), and public datasets like \climaText and \climateFEVER. In this paper, we evaluate and compare the performance of ClimateBERT on diverse CC tasks for the first time, providing a comprehensive, publicly available and reproducible evaluation.

\begin{table}[!ht]
\centering
\resizebox{0.6\columnwidth}{!}{%
\begin{tabular}{ll}
\hline
\textbf{Parameter} & \textbf{Values}       \\ \hline
loss               & hinge, squared\_hinge \\ \hline
C                  & 0.01, 0.1, 1, 10      \\ \hline
class\_weight      & none, balanced        \\ \hline
\end{tabular}
}
\caption{For the linear SVM, we grid search over the parameters with 5-fold validation to get the best fit out of 80 candidates (16 values * 5 folds) with F1 Macro as the scoring mechanism}
\label{tab:svm_params}
\end{table}
\begin{SCtable*}[1]
\centering
\setlength\extrarowheight{7pt}
\resizebox{.7\textwidth}{!}{%
\begin{tabular}{@{}rrrrrrrl@{}}
\toprule
\textbf{} &
  \multicolumn{2}{|c|}{\textbf{\cdpCities}} &
  \multicolumn{2}{c|}{\textbf{\cdpStates}} &
  \multicolumn{2}{c}{\textbf{\cdpCorps}} &
   \\ \midrule
\multicolumn{1}{r|}{\textbf{Model}} &
  \multicolumn{1}{r|}{\textbf{MRR@10}} &
  \multicolumn{1}{r|}{\textbf{MRR@All}} &
  \multicolumn{1}{r|}{\textbf{MRR@10}} &
  \multicolumn{1}{r|}{\textbf{MRR@All}} &
  \multicolumn{1}{r|}{\textbf{MRR@10}} &
  \textbf{MRR@All} &
   \\ \midrule
\multicolumn{7}{c}{\textbf{No Finetuning on \cdp}} &
   \\ \midrule
\multicolumn{1}{r|}{\textbf{BM25}} &
  \multicolumn{1}{r|}{0.055} &
  \multicolumn{1}{r|}{0.077} &
  \multicolumn{1}{r|}{0.084} &
  \multicolumn{1}{r|}{0.105} &
  \multicolumn{1}{r|}{0.153} &
  0.180 &
   \\ \midrule
\multicolumn{1}{r|}{\textbf{MiniLM}} &
  \multicolumn{1}{r|}{0.099} &
  \multicolumn{1}{r|}{0.118} &
  \multicolumn{1}{r|}{0.120} &
  \multicolumn{1}{r|}{0.142} &
  \multicolumn{1}{r|}{0.320} &
  0.342 &
   \\ \midrule
\multicolumn{7}{c}{\textbf{Finetuned on \cdp}} &
   \\ \midrule
\multicolumn{1}{r|}{\textbf{}} &
  \multicolumn{2}{c|}{In-Domain} &
  \multicolumn{2}{c|}{In-Domain} &
  \multicolumn{2}{c}{In-Domain} &
   \\ \midrule
\multicolumn{1}{r|}{\textbf{ClimateBERT}} &
  \multicolumn{1}{r|}{0.331} &
  \multicolumn{1}{r|}{0.344} &
  \multicolumn{1}{r|}{0.422} &
  \multicolumn{1}{r|}{0.431} &
  \multicolumn{1}{r|}{0.753} &
  0.754 &
   \\ \midrule
\multicolumn{1}{r|}{\textbf{MiniLM}} &
  \multicolumn{1}{r|}{\textbf{0.366}} &
  \multicolumn{1}{r|}{\textbf{0.378}} &
  \multicolumn{1}{r|}{0.482} &
  \multicolumn{1}{r|}{0.491} &
  \multicolumn{1}{r|}{\textbf{0.755}} &
  \textbf{0.757} &
   \\ \midrule
\multicolumn{7}{c}{\textbf{Best Model Finetuned on all}} &
   \\ \midrule
\multicolumn{1}{r|}{\textbf{MiniLM}} &
  \multicolumn{1}{r|}{0.352} &
  \multicolumn{1}{r|}{0.364} &
  \multicolumn{1}{r|}{\textbf{0.489}} &
  \multicolumn{1}{r|}{\textbf{0.497}} &
  \multicolumn{1}{r|}{0.745} &
  {0.747} &
   \\ \bottomrule
\end{tabular}%
}
\caption{MRR@$k$ scores for BM25, ClimateBERT and MSMARCO-MiniLM on the three subsets of \climaQA. Models finetuned and evaluated on same subset fall under In-Domain.}
\label{tab:mrr_scores_full}
\end{SCtable*}

\begin{table}[!ht]
\centering
\resizebox{\columnwidth}{!}{%
\begin{tabular}{@{}rrrrrr@{}}
\toprule
\multicolumn{1}{r|}{\textbf{}} &
  \multicolumn{2}{c|}{\textbf{\cdpStates}} &
  \multicolumn{2}{c}{\textbf{\cdpCorps}} \\ \midrule
\multicolumn{1}{r|}{\textbf{Model}} &
  \multicolumn{1}{r|}{\textbf{MRR@10}} &
  \multicolumn{1}{r|}{\textbf{MRR@All}} &
  \multicolumn{1}{r|}{\textbf{MRR@10}} &
  \textbf{MRR@All} \\ \midrule
\multicolumn{5}{c}{\textbf{No Finetuning}} \\ \midrule
\multicolumn{1}{r|}{\textbf{BM25}} &
  \multicolumn{1}{r|}{0.084} &
  \multicolumn{1}{r|}{0.105} &
  \multicolumn{1}{r|}{0.153} &
  0.180 \\ \midrule
\multicolumn{1}{r|}{\textbf{MiniLM}} &
  \multicolumn{1}{r|}{0.120} &
  \multicolumn{1}{r|}{0.142} &
  \multicolumn{1}{r|}{0.320} &
  0.342 \\ \midrule
\multicolumn{5}{c}{\textbf{Finetuned on \cdpCities}} \\ \midrule
\multicolumn{1}{r|}{\textbf{}} &
  \multicolumn{2}{c|}{Transfer} &
  \multicolumn{2}{c}{Transfer} \\ \midrule
\multicolumn{1}{r|}{\textbf{ClimateBERT}} &
  \multicolumn{1}{r|}{0.298} &
  \multicolumn{1}{r|}{0.314} &
  \multicolumn{1}{r|}{0.465} &
  0.477 \\ \midrule
\multicolumn{1}{r|}{\textbf{MiniLM}} &
  \multicolumn{1}{r|}{\textbf{0.353}} &
  \multicolumn{1}{r|}{\textbf{0.366}} &
  \multicolumn{1}{r|}{\textbf{0.489}} &
  \textbf{0.500} \\ \bottomrule
\end{tabular}%
}
\caption{MRR@k scores for BM25, ClimateBERT and MSMARCO-MiniLM on the Transfer experiments. Models are finetuned on \cdpCities and evaluated on States and Corporations.}
\label{tab:mrr_scores_transfer_full}
\end{table}

\begin{table*}[ht]
\resizebox{2\columnwidth}{!}{
\begin{tabular}{p{14cm}HHr}
\toprule
\textbf{Question}                                                                                                                                                                                                                                                                                                                                  & \multicolumn{1}{H}{\textbf{no. of samples}} & \multicolumn{1}{H}{\textbf{NDCG@132}} & \multicolumn{1}{l}{\textbf{MRR@132}} \\ \midrule
Please provide details of your climate actions in the Agriculture sector.                                                                                                                                                                                                                                                                          & 29.00                                       & 0.900                                 & 0.870                                \\
Please provide details of your climate actions in the Waste sector.                                                                                                                                                                                                                                                                                & 58.00                                       & 0.833                                 & 0.789                                \\
Please provide details of your climate actions in the Transport sector.                                                                                                                                                                                                                                                                            & 49.00                                       & 0.819                                 & 0.774                                \\
Please provide details of your climate actions in the Buildings \& Lighting sector.                                                                                                                                                                                                                                                                & 42.00                                       & 0.687                                 & 0.597                                \\
Please describe these current and/or anticipated impacts of climate change.                                                                                                                                                                                                                                                                        & 61.00                                       & 0.613                                 & 0.492                                \\
Please complete the table below.                                                                                                                                                                                                                                                                                                                   & 62.00                                       & 0.602                                 & 0.487                                \\
Please indicate the opportunities and describe how the region is positioning itself to take advantage of them.                                                                                                                                                                                                                                     & 23.00                                       & 0.577                                 & 0.445                                \\
Please provide details of your climate actions in the Energy sector.                                                                                                                                                                                                                                                                               & 55.00                                       & 0.538                                 & 0.397                                \\
Please describe the adaptation actions you are taking to reduce the vulnerability of your region's citizens, businesses and infrastructure to the impacts of climate change identified in 6.6a.                                                                                                                                                    & 59.00                                       & 0.517                                 & 0.378                                \\
Please describe these current and/or future risks due to climate change.                                                                                                                                                                                                                                                                           & 27.00                                       & 0.483                                 & 0.327                                \\
List any emission reduction, adaptation, water related or resilience projects that you have planned within your region for which you hope to attract financing, and provide details on the estimated costs and status of the project. If your region does not have any relevant projects, please select “No relevant projects” under Project Area. & 20.00                                       & 0.473                                 & 0.319                                \\
Please provide details of your climate actions in the  Land use sector.                                                                                                                                                                                                                                                                            & 27.00                                       & 0.447                                 & 0.286                                \\
Please provide the details of your region-wide base year emissions reduction target(s). You may add rows to provide the details of your sector-specific targets by selecting the relevant sector in the sector field.                                                                                                                              & 27.00                                       & 0.423                                 & 0.252                                \\
Please describe the adaptation actions you are taking to reduce the vulnerability of your region's citizens, businesses and infrastructure to the risks due to climate change identified in 5.4a.                                                                                                                                                  & 23.00                                       & 0.414                                 & 0.247                                \\ \bottomrule
\end{tabular}}
\caption{Question difficulty evaluated on the test set of \cdpStates ranked from best performing to worst performing. Filtered to only questions that appeared at least twenty times.}
\label{tab:qs-dif}
\end{table*}

\section{Climate Text Sources}\label{app:more-info}
The reports considered here include climate assessments, climate legislation, agency reports, regulatory filings, climate action plans (CAPs), and corporate ESG (Environmental, Social, and Governance) and CSR (Corporate Social Responsibility) documents~\cite{esg,csr}.

A key step in curbing emissions and mitigating climate change has been the development of standards and frameworks for climate reporting such as GRI (Global Reporting Initiative), TCFD (Task Force on Climate-Related Financial Disclosures), CDP (Carbon Disclosure Project), SASB (Sustainability Accounting Standards Board), and SDG (Sustainable Development Goals)~\cite{gri,tcfd,cdp,sdg}

For example, the World Resources Institute has built Climate Watch~\cite{climatewatch} to keep track of progress and commitments nations have made under the 2015 Paris Agreement. One element of their work has been the manual labeling of Nationally Determined Contributions (NDCs) with a number of descriptors including cross references to the UN Sustainable Development Goals which strongly overlap with the categories in our CDP dataset and task.

Although SDGs were first established by the United Nations to measure the progress of nation states towards development goals, they have been adopted by both corporations and regional and local jurisdictions to measure their sustainability efforts.

However, since the cross-referencing with SDGs is largely voluntary many cities, for example, have CAPs that are hundreds of pages in length but that provide no alignment with SDGs. Being able to effectively align the text between different climate documents to the various standards and disclosure frameworks is a critical component of climate policy evaluation and a real-world challenge. See also ~\citet{stede-patz-2021-climate} for more in-depth information.

\begin{table*}[t!]
    \resizebox{2\columnwidth}{!}{
    \begin{tabular}{lllp{11cm}}
    \toprule
    Ex. &
       &
      \textbf{Text Segment from State Climate Action Plans} &
      \textbf{Top Questions from \cdpStates (using fine-tuned MiniLM)} \\ \midrule
    
    \multirow{2}{*}{1} &
      &
      \multirow{2}{11cm}{By a majority vote, the ICCAC presents a policy option that, if deemed necessary, would build one new 1200-megawatt nuclear power plant in Iowa by January 1, 2020.} &
      Q3: Please provide details of your renewable energy or electricity target(s). \\
     &
       &
      &
      Q4: Please provide details of your climate actions in the Energy sector. \\ \midrule
    
    \multirow{2}{*}{2} &
      &
      \multirow{2}{11cm}{California maintains a GHG inventory that is consistent with IPCC practices ... Reports from facilities and entities that emit more than 25,000 MTCO2e are verified by a CARB-accredited third-party verification body.} &
      Q1: Please give the name of the primary protocol, standard, or methodology you have used to calculate your government’s GHG emissions. \\
     &
       &
      &
      Q3: Please provide the following information about the emissions verification process. \\ \midrule
    
    \multirow{2}{*}{3} &
       &
      \multirow{2}{11cm}{A leading driver of these high emissions is the fact that the District’s daytime population swells by 400,000 workers every workday, which is the largest percentage increase in daytime population of any large city in the nation.} &
      Q4: Please indicate if your region-wide emissions have increased, decreased, or stayed the same since your last emissions inventory, and please describe why. \\
     &
       &
      &
      Q5: Please report your region-wide base year emissions in the table below. \\ \midrule
    
    \hline
    \end{tabular}}
    \caption{More examples from our human pilot study.}\label{tab:cherry2}
    \end{table*}

\end{document}